\documentclass[a4paper,12pt] {article}
\usepackage[english]{babel}
\usepackage[utf8]{inputenc}
\usepackage{authblk}
\usepackage{graphicx}
\usepackage{hyperref}
\usepackage{dirtytalk}
\usepackage{threeparttable}
\usepackage{apacite}

\title{\large{\textbf{MICE: A Crosslinguistic Emotion Corpus \\ in Malay, Indonesian, Chinese and English\footnote{For related publications, please see Ng, Cui, Cavallaro (2019), Susanto, Livingstone, Ng, and Cambria (2020), Susanto and Ng (2021a), Susanto and Ng (2021b), Yap (2021).}}}} 
\author{\small{Ng Bee Chin, Yosephine Susanto, Erik Cambria}}
\affil{\vspace{-2ex}Nanyang Technological University, Singapore}

\date{}

\begin{document}

\maketitle
\vspace{-5ex}
\begin{abstract}
MICE is a corpus of emotion words in four languages which is currently working progress. There are two sections to this study – Part I:  Emotion word corpus and Part II: Emotion word survey. In Part 1, the method of how the emotion data is culled for each of the four languages will be described and very preliminary data will be presented. In total, we identified 3,750 emotion expressions in Malay, 6,657 in Indonesian, 3,347 in Mandarin Chinese and 8,683 in English. We are currently evaluating and double checking the corpus and doing further analysis on the distribution of these emotion expressions. Part II Emotion word survey involved an online language survey which collected information on how speakers assigned the emotion words into basic emotion categories, the rating for valence and intensity as well as biographical information of all the respondents. 
\end{abstract}

\section{Introduction}
Emotion research, sometimes referred to as ‘Affective Science’ is a burgeoning field attracting different lines of enquires from multiple disciplines like psychology, anthropology and computer science to name a few. This interdisciplinary interest in emotion is not surprising given its indisputable and vital role in human interaction. Despite this, linguists have only entered the discussion in the last 25 years. This is because emotion has traditionally been perceived as a mental state, independent of language. In the last 15-20 years however, many scientists are revisiting the Sapir-Whorf concept that language constrains our interpretations of events and cognition \cite{pavlenko2014bilingual}. This presents the scientific community with a challenging field of inquiry into cultural aspects of expressing affect in diverse communicative settings. However, though there is a surge in interest and studies on language and emotion, there is no agreement on the representations and classification of emotions across languages and cultures to allow for a coherent comparison. Thus, this proposed study aims to systemise classificatory issues in 3 prominent languages spoken in Singapore, namely English, Mandarin Chinese and Malay, as well as an important neighbouring language, Indonesian by establishing a comparable corpus (abbreviated as MICE). In doing so, we hope that the data will provide a starting point for an overall comparison of emotion terms. This MICE corpus will allow for comparative work on the four crucial languages spoken in this region. Apart from the frequency and prevalence of references to emotion, the corpus will be a significant stepping stone for different avenues of inquiry including intercultural research, psychological assessments, artificial intelligent systems and sentiment analysis. From this common database, we will be able to develop multiple language and emotion studies to gain important insights into the impact of emotion on communication.

\section{Language and Emotion}
A vast number of approaches to the study of emotion and various theories have been proposed to date, with naturalism and social constructionism being two major positions adopted in the academic community \cite{ogarkova2007green, pavlenko2014bilingual}. Traditional studies originally focused on the naturalist approach by determining universal states of emotion without linguistic or cultural considerations. However, the common problem encountered by this approach is determining how many basic emotional states exist, which researchers have failed to agree upon \cite{pavlenko2014bilingual}. Though \citeA{ekman2003emotions} seven basic emotions (\textit{anger, disgust, contempt, surprise, fear, happiness} and \textit{sadness}) have been put forth as universal, there has been much debate about the validity of this claim. More recently, \citeA{jack2014dynamic} proposed that there are only four basic emotions – \textit{glad, sad, mad} and \textit{scared}. Attesting the universality of emotion has always been an issue with researchers working from a ‘universalist’ perspective. Furthermore, it is difficult, if not impossible, to dissociate the study of emotion from language and many of the early researchers have been criticised for their eurocentric approach to the field. \par
The language used in academic writing, particularly English, is the technical medium used to describe research findings on emotion studies and as aptly pointed out by \citeA{enfield2002introduction}, English is after all, not a culturally neutral language. They argued that using English as a metalanguage to discuss emotions, especially in other languages, would obscure diversity in favour of universalist interpretations. Furthermore, it is clear that not all languages encode emotional meaning in the same manner, as different studies have found variances in emotion words across languages. There is a huge range varying between 7 (e.g. Chewong, \citeNP{howell1981rules}) to 2000 (e.g. English, \citeNP{wallace1973sharing}). One suspects that categorisations and enumeration of emotion or emotion words largely depends on the set of criteria used. The primary goal of this study is to apply the same criteria across different languages so that there is a common basis for comparison.

\section{Part I: Emotion Corpus}
Given the complexities of emotion cross- linguistically and culturally, there is a need for a systematic approach to its study. \citeA{scherer2013measuring} advocated taking a lexical approach, suggesting that, given the primacy of emotion in social interaction, adequate and meaningful representations of emotion states must be present in the substance of language. Examining the semantic organisation and distribution of emotion words can shed light on the structure of emotion concepts, revealing universal as well as culture- and language-specific features. \par
There are several existing emotion wordlists such as the Affective Norms for English Words (ANEW; \citeNP{bradley1999affective}). This was designed to serve as stimuli databases and do not represent the full working emotion vocabularies of a given language. Additionally, these word lists are difficult to compare cross-linguistically as they vary widely in their manner of creation and characterisation, stemming from the different emotion models upon which they were based. \par
There were others but they all used varying semantic dimensions to specifically distinguish emotion from non-emotion words and also used different combinations (e.g. valence, intensity, and duration in \citeNP{Zammuner998concepts} and valence, arousal, and imageability in \citeNP{whissell2008comparison}). Other word lists are instead informed by discrete emotion models. In \citeA{DOOST1999development}'s study, emotion words were elicited from participants imagining and describing a basic emotion state. \citeA{stevenson2007characterization} and \citeA{briesemeister2011discrete} had participants rate and characterise ANEW words into five predetermined discrete emotion categories - \textit{happiness, sadness, anger, fear} and \textit{disgust}. As such, the different methodologies for studying emotion and language make it difficult, or nearly impossible, to make reliable comparisons between studies and more so when working across different languages. \par
The creation of emotion corpora is also hampered by a lack of standardisation in identifying emotion words. While several approaches have been used (e.g. \citeNP{ortony1987referential}; \citeNP{johnson1989language}), it is debatable as to which method is most suitable or appropriate. One possibility involves the propositional analysis adopted by \citeA{wallace1973sharing}, by extracting adjectives and nouns that fit the syntactic contexts \say{He has a feeling of X} and \say{He feels X}. A main issue is that such an approach elicits a host of emotion words which may differ qualitatively in the way they relate to emotion. For example, ‘sadness’, ‘cry’ and ‘idiot’ have all been categorized as emotion words but as pointed out by \citeA{pavlenko2008emotion}, they are qualitatively different in the emotion palette. While `sadness’ is an emotion word, `cry’ is related to an emotion – and this emotion could be sadness, joy, fear, shame, etc. Hence, it is termed ‘emotion-related’. `Idiot’ on the other hand, evokes an emotion from the interlocutor, as do `darling’, `shit’ or other swearwords or endearments. This type of emotion provoking words is classified as ‘emotion laden’ words. \citeauthor{pavlenko2008emotion} argued strongly that `emotion words’, `emotion laden words’ and `emotion-related words’ be given careful attention. To date, most studies have not really taken this into account and the impact of this semantic variation on cognition is only just beginning to be empirically validated \cite{altarriba2010representation, sutton2016finding}.\par
Moreover, the propositional approach used in previous studies is language-specific. The linguistic structures commonly used to express emotion in one language may not be applicable to others. In English and French, emotions are usually expressed using adjectives, while in Russian or Polish emotions are more frequently described using verbs referring to processes and relationships \cite{pavlenko2002emotions,wierzbicka1992defining,wierzbicka2004preface}. Many assumptions are thus made with regards to the type of emotions to explore and the working definitions of what can be considered as emotion words. Consequently, after 15 years of explorations, the field of language and emotion is still shrouded in uncertainties. Though pairwise language comparisons in specific domains or with specific emotion words have yielded fascinating results, we are still far from seeing the complete landscape. Thus far, we have spotlights illuminating specific corners, and what is really needed is the opportunity to have a overview of emotion terms of languages holistically.

\section{Preliminary/Previous Studies}
The researchers of this current study built a corpus of Mandarin Chinese emotion words in an attempt to address the limitations of past studies on emotion corpora. In the study, \citeA{Ng2016semantics} used a standard Mandarin Chinese dictionary to extract and categorise emotion words. A dictionary was used instead of a natural corpus so that coverage could be comprehensive. The words were identified based on the parameters discussed in \citeA{pavlenko2008emotion}. The identified words were also coded for emotion word types, emotion domains, valence, intensity, part-of-speech and frequency of use.\par
The results of their study provided evidence of a meaningful framework for classifying and coding emotion words in Mandarin Chinese. Specifically, they were able to statistically distinguish three emotion word categories; emotion words, emotion related words and emotion-laden words. The data also appeared to support the idea of a distinct prevalence of emotion words which are verbs in Chinese, and this is significantly more pronounced in emotion-related words. Hence, just at first glance, the corpus strikes at the heart of the study of emotion lexicon. Up till now, researchers have drawn randomly from a set of emotion words which have not been semantically analysed and if they were analysed, there were no systematic methods adopted across studies. The viability of the framework in this Mandarin Chinese study thus helps to set the groundwork for this proposed project. It will also help to provide comparable data from other languages to verify the claim that verbs are more prevalent in Mandarin Chinese.

\section{Specific Aims}
MICE emotion corpus built for the purpose of providing a comparison emotion lexicon across four different languages: Malay, Indonesian, Chinese, English allows the users to:
\begin{itemize}
    \item{Explore the language specificity in the semantic organisation and distribution of emotion words in a language.}
    \item{Develop an intercultural understanding of emotion through comparisons of emotion corpora studies.}
    \item{Enable the corpus to be used by computational linguists for the purpose of sentiment analysis.}
\end{itemize}

\section{Significance}
\subsection{Build a Corpus – filling a needed gap}
In the field of language and emotion research – the first thing researchers ask for is a list or a set of emotion words related to `X’ emotion. What are expressions of `anger’, `pride’, `shame’, or `guilt’? Up to now, various studies have been done with researchers using their own lists of emotion terms. Hence, in comparative work, we are often not comparing similar measures. Though there is a dire need for a comprehensive corpus, none has been done as the process is tedious and so far, the criteria for determining emotion words have been controversial. The Mandarin Chinese corpus is already done, and we can now focus on Malay, Indonesian and English. The corpus will be a valuable resource for language and emotion researchers to access and extract the information they need. In addition, it will allow us to address the absence of an empirical and unified framework for studying emotions across languages. This project will also help establish a precedent for future research by creating a standardised corpus database. 

\subsection{Facilitate research on language comparison}
The corpus once completed, will be able to facilitate many other studies on language and emotion. The focus on these 3 core languages used regionally, English, Mandarin Chinese, Malay and Indonesian will allow for future intercultural studies that have potential impact on improving communication in the region. An Indonesian colleague, Dr Dwi Noverini Djenar from the University of Sydney will help in analysing the Indonesian database using \citeA{Ng2016semantics} approach. Given the nuanced differences in emotion and its potential to affect communication – sometimes with deleterious outcomes, this parallel development of Malay and Indonesian corpus has strong significance for the region. Though Malay and Indonesian are mutually intelligible, there are numerous and significant communicative differences when it comes to emotion. For example, \textit{berang} in Indonesian can mean `angry’, while in Malay, it means `sea otter’. Other examples include, \textit{galau} meaning `uncertain, indecisive’ in Indonesian but in Malay it means `commotion’. Indonesian also has many loanwords from Javanese including emotion terms (e.g. \textit{kangen} `to miss someone’; \textit{demen} `to like someone or something, fond of’; \textit{mangkel} `annoyed’; \textit{ngambek} `sulking’). These are not used in Malay. Given that MICE are the main languages in this region, the comparison will yield important findings for ensuring more effective intercultural communication.

\subsection{Computational application}
The emotion corpora developed can also provide us with a potential database for use in future research to address other academic questions. Up to now, sentiment analysis has relied on measures of valence and intensity to extract information. The advantage of sentiment analysis is its ability to process huge sets of data and draw conclusions based on numerical strengths. However, the cost of this approach is the loss of finer grained meaning distinctions, an inherent quality of human language. A corpus built by a team of linguists will have more capacity to tackle meaning distinctions that elude machines. In turn, computational linguists such as Erik Cambria, the Co-PI in this proposal, and his research team (\url{http://sentic.net/team}) will be able to incorporate his experience in computational approaches to help with building an efficient corpus that will facilitate their sentiment analysis research \cite{cambria2017affective}. \citeA{poria2015sentiment}, in fact, have advocated for the importance of incorporating affective information into a common-sense reasoning framework for sentiment analysis. The availability of such an emotion corpus, hence, could have potential benefits for business management and strategies. Emotion corpora would be able to provide the tools, i.e. linguistic information and database of emotion words, to complement the process of sentiment analysis in different languages. Hence, there is a potential opportunity for the current project to contribute to research in a multi-disciplinary manner, and also to promote multi-disciplinary collaborations. 

\section{Methodology for Part I}
A team of research assistants  (8-18) who are speakers of each language are trained to recognise emotion expressions and they were also trained to categorise such emotion expressions into the three categories of emotion words, emotion related words and emotion-laden words identified by \citeA{pavlenko2008emotion}. This training is done in a workshop and the research assistants worked in pairs on the same list of words. In step 1, the words are culled from the most commonly used dictionaries in each language. The dictionary is divided approximately into 10 sections and each pair of research assistants start with 1 section of the dictionary. Next, the RAs were asked to compare their list and every discrepancy was discussed and resolved together in the team.\par

\begin{figure}[ht]
    \centering
    \includegraphics[width=0.8\textwidth]{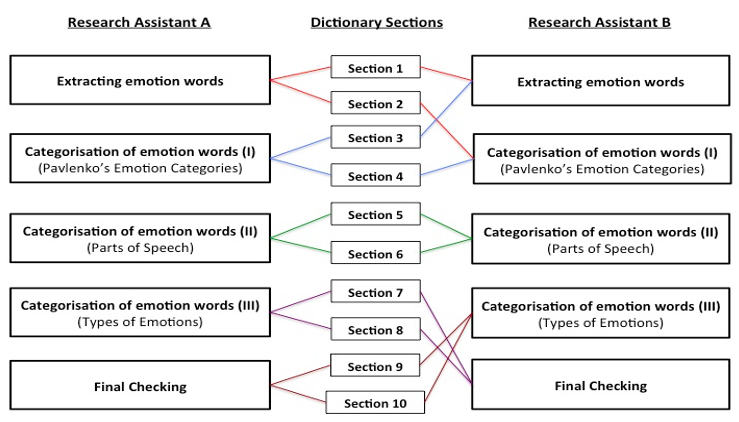}
    \caption{Example of how two RAs may be assigned the 10 sections for different assessments}
    \label{fig1}
\end{figure}

\begin{figure}[!ht]
    \centering
    \includegraphics[width=0.9\textwidth]{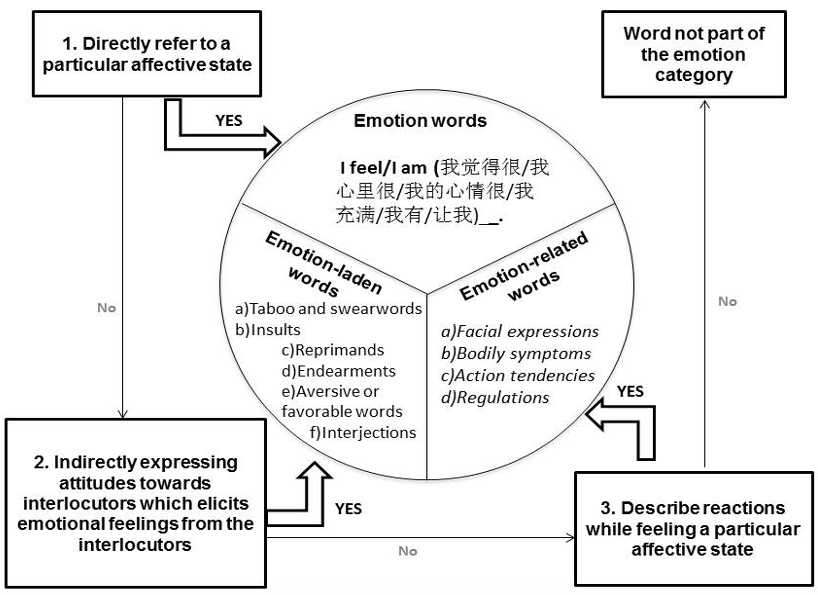}
    \caption{Sample framework for classifying emotion words \protect\cite{Ng2016semantics}}
    \label{fig2}
\end{figure}

\newpage
Step 2 involves the categorising of the emotion words into the three categories of emotion words, emotion-related words and emotion-laden words. Again, any discrepancy was discussed and resolved.   Step 3 involved classifying the words into the parts of speech. This process takes about 2-3 weeks of full-time work by each language team. Figure \ref{fig1} presents the flow of work described above. Figure \ref{fig2} presents the culling process and the decision criteria for the categorisation. In general, all research assistants work in pair until they achieve 100\% agreement on the sections they were assigned. The results for the Chinese emotion lexicon is available and for more details on the process, please see \citeA{ng2019annotated} for more details. \par

The following is a brief report of the data we have now (Table \ref{tab1}). 

\begin{table}[hb]
\centering
\caption{Emotion expressions in MICE}
\label{tab1}
\resizebox{\textwidth}{!}
{%
\begin{tabular}{lllll}
           & Emotion words & Emotion related words & Emotion laden words & Total \\ \hline
Malay      & 554           & 1,223                 & 1,973               & 3,750 \\
Indonesian & 609           & 721                   & 5,327               & 6,657 \\
Chinese    & 953           & 978                   & 1,416               & 3,347 \\
English    & 665           & 1,487                 & 6,531               & 8,683
\end{tabular}%
}
\end{table}

\newpage
\section{Part II: Emotion Survey}
In part two of the study, we aim to  compare the range and depth of emotion terms across the four languages. We ask native speakers of the studied languages to rate the valence and intensity of  20-30 emotion words  by using Qualtrics online questionnaire in a single session. Participants were also asked to assign each of the emotion word into one of the 7 basic emotion categories. This study is based in Singapore and Indonesia. Biographical data and language use patterns were also gathered to look at possible variables affecting the valence and intensity ratings. The target survey response rate is 2000 for each language. So far, only Indonesian and English have met this target. Data collection for the other languages are still ongoing.  The current sample size for each language can be seen in Table \ref{tab2} below.\par

\begin{table}[ht]
\centering
\caption{Number of survey responses for each language}
\begin{threeparttable}
\label{tab2}
{%
\begin{tabular}{ll}
           & Participants \\ \hline
Malay\tnote{*}     & NA           \\
Indonesian & 3,216        \\
Chinese    & 1,226        \\
English    & 2,128       

\end{tabular}%
}
\begin{tablenotes}
\item[*]{\footnotesize{Note: upcoming}}
\end{tablenotes}
\end{threeparttable}
\end{table}

This survey will be very beneficial for emotion research, especially intercultural research, psychological assessments, artificial intelligent systems and sentiment analysis. From this multilingual database, we will able to develop multiple language and emotion studies to gain important sights into the impact of emotion on communication.

\section*{Acknowledgements}
The development of the corpus is supported by a AcRF Tier 1 funding (MICE - A Multilingual Corpus of Emotion Expressions of Malay, Indonesian, Chinese and English (000663-00001), Ng Bee Chin (PI) and Erik Cambria (Co-PI), Grant period: 01/03/2017-31/08/2021).

\nocite{ng2019annotated}
\nocite{susanto2020hourglass}
\nocite{Susanto2020kolita}
\nocite{Susanto2021isloj}
\nocite{Susanto2021ismil}
\nocite{Yap2021mood}

\bibliographystyle{apacite}
\bibliography{references}

\end{document}